\documentclass[letterpaper,10 pt, conference]{ieeeconf}
\pdfoutput=1
\IEEEoverridecommandlockouts

\usepackage{amsmath}
\usepackage{amssymb}
\usepackage{mathtools}
\usepackage{color}
\usepackage{graphicx}
\usepackage{caption}
\usepackage{subcaption}
\usepackage{times}
\usepackage{times}
\usepackage{epsfig}
\usepackage{bm}
\usepackage[noadjust]{cite}
\newtheorem{definition}{Definition}

\newcommand{\V}[1]{\boldsymbol{#1}}

\renewcommand{\baselinestretch}{0.84}

\title{\LARGE \bf A Framework for Robot Manipulation: \\ Skill Formalism, Meta Learning and Adaptive Control}

\author{Lars Johannsmeier, Malkin Gerchow and Sami Haddadin$^{1}$
\thanks{$^{1}$Authors are with Chair of Robotic Sciences and System Intelligence,
        Technische Universit\"at M\"unchen, 81797 M\"unchen, Germany
        {\tt\small lastname@irt.uni-hannover.de}}
}

\begin{document}

\maketitle
\thispagestyle{empty}
\pagestyle{empty}

\begin{abstract}
In this paper we introduce a novel framework for expressing and learning force-sensitive robot manipulation skills. It is based on a formalism that extends our previous work on adaptive impedance control with meta parameter learning and compatible skill specifications. This way the system is also able to make use of abstract expert knowledge by incorporating process descriptions and quality evaluation metrics. We evaluate various state-of-the-art schemes for the meta parameter learning and experimentally compare selected ones. Our results clearly indicate that the combination of our adaptive impedance controller with a carefully defined skill formalism significantly reduces the complexity of manipulation tasks even for learning  peg-in-hole with submillimeter industrial tolerances. Overall, the considered system is able to learn variations of this skill in under 20 minutes. In fact, experimentally the system was able to perform the learned tasks faster than humans, leading to the first learning-based solution of complex assembly at such real-world performance.
\end{abstract}

\section{Introduction}\label{sec:introduction}
Typically, robot manipulation skills are introduced as more or less formal representations of certain sets of predefined actions or movements. Already, there exist several approaches to programming with skills, e.g. \cite{pedersen2015robot,thomas2013new,andersen2014definition}. A common drawback is, however, the need for laborious and complex parameterization resulting in a manual tuning phase to find satisfactory parameters for a specific skill.
Depending on the particular situation various parameters need to be adapted in order to account for different environment properties such as rougher surfaces or different masses of involved objects. Within given boundaries of certainty they could be chosen such that the skill is fulfilled optimally with respect to a specific cost function. This cost function and additional constraints are usually defined by human experts optimizing e.g. for low contact forces, short execution time or low power consumption. Typically, manual parameter tuning is a very laborious task, thus autonomous parameter selection without complex pre-knowledge about the task other than the task specification and the robot prior abilities is highly desirable. However, such automatic tuning of control and other task parameters  in order to find feasible, ideally even optimal parameters, in the sense of a cost function is still a significant open problem in robot skill design

\begin{figure}[ht!]
\begin{center}
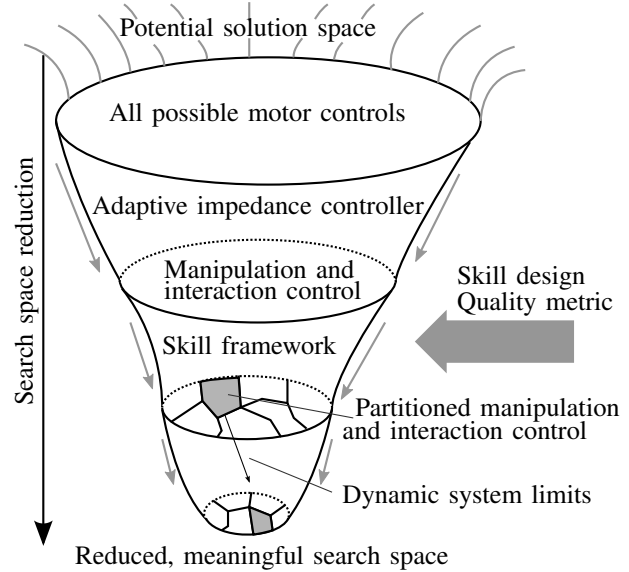
\caption{Meaningful complexity reduction of the search space by applying adaptive impedance control, a skill formalism and inherent system limits.}
\label{fig:complexity}
\end{center}
\vspace{-0.8cm}
\end{figure}

So far, several approaches were proposed to tackle this problem. In \cite{pastor2009learning}, e.g. learning motor skills by demonstration is described. In \cite{pastor2011skill} a Reinforcement Learning (RL) based approach for acquiring new motor skills from demonstration is introduced. \cite{kober2009learning,kober2009policy} employ RL methods to learn motor primitives that represent a skill. In \cite{schaal2005learning} supervised learning by demonstration is used in combination with dynamic movement primitives (DMP) to learn bipedal walking in simulation. An early approach utilizing a stochastic real-valued RL algorithm in combination with a nonlinear multilayer artificial neural network for learning robotic skills can be found in \cite{gullapalli1994acquiring}. In \cite{levine2015learning}, guided policy search was applied to learn various manipulation tasks based on a neural network policy. 
The drawbacks of many existing approaches are their high demand for computational power and memory, e.g. in form of GPUs and computing clusters, and with a few exceptions they require a large amount of learning time already in simulation.

\renewcommand{\arraystretch}{1.3}
\begin{table*}[ht!]
\begin{center}
\caption{Overview of existing work about peg-in-hole}
\label{tab:ref_peginhole}
\begin{tabular}{|p{1.2cm}|p{3.5cm}|p{2cm}|p{1.7cm}|p{1.5cm}|p{2.5cm}|p{2.5cm}|}
\hline
Reference & Principle & Motion strategy & Underlying controller &  Adaptation / Learning principle & Learning / Insertion speed & Difficulty \\
\hline
\cite{chhatpar2001search,broenink1996peg,lozano1984automatic,bruyninckx1995peg,newman2001interpretation} & Geometric approaches, analysis of peg-in-hole and force/moment guidance & Reference trajectory  & Impedance, admittance, force control & none & Insertion times vary between $5$-$11$ s & Round pegs with tolerances between $0.25$-$1$ mm.\\
\hline
\cite{stemmer2006robust} & Combination of visual perception and compliant manipulation strategies to approach a difficult peg-in-hole problem & Trajectory generator & Impedance control & none & Insertion time was about $2$-$3$ s
 & Multiple pieces of different forms were used with tolerances of about $0.1$ mm. \\
\hline
\cite{gullapalli1992learning,gullapalli1994acquiring} & Peg-in-hole skill is learned via reinforcement learning. & Policy in terms of trajectory is represented by neural networks & Admittance, force control & Reinforcement learning & Learning took about $500$-$800$ trials, insertion time is between $20$-$40$ s after learning & 2D-problem with tolerance of $0.127$ mm in simulation and a round peg with a tolerance of $0.175$ mm in experiments.)\\
\hline
\cite{nemec2013transfer,kramberger2016transfer} & DMPs are adapted based on measurements of the external wrench to adapt demonstrated solutions of a peg-in-hole task  & Trajectory from DMP & Impedance, Admittance, Force & Trajectory adaptation &  Adaptation requires about $5$ cycles. Insertion requires about $8$-$10$ s based on human demonstration. & Several basic shapes and and varying tolerances of $0.1$-$1$ mm. \\
\hline
\cite{kronander2014task}& Several controllers (with and without adaptation) were compared for a peg-in-hole task. The initial strategy is learned from human demonstration & (Initial) reference trajectory & Impedance control & Adaptation of orientation via random sampling and Gaussian mixture models& Average insertion times of $6$-$9$ s were achieved, $50$ trials were used for sampling. Data acquisition by human demonstration initially is necessary. & Round peg and hole made of steel and a wooden peg with a rubber hole were used.\\
\hline
\cite{levine2015learning,levine2016end,devin2017learning}& Policies represented by convolutional neural networks are learned via reinforcement learning based on joint states and visual feedback as an end-to-end solution. Among other tasks peg-in-hole has been learned mostly in simulation but also in real world experiments. & Policies directly generate joint torques & linear-Gaussian controllers / neural network control policies & Reinforcement learning & Depending of the task learning took $100$-$300$ trials in simulation and real-world experiments. Additional computational requirements of about $50$ minutes are mentioned & Tasks such as bottle opening and simpler variations of peg-in-hole were learned.\\
\hline
\end{tabular}
\end{center}
\vspace{-0.6cm}
\end{table*}

In order to execute complex manipulation tasks such as inserting a peg into a hole soft robotics controlled systems like Franka Emika's Panda \cite{haddadin2017franka} or the LWR III \cite{hirzinger2002dlr} are the system class of choice.
Early work in learning impedance control, the most well-known soft-robotics control concept, is for example introduced in \cite{678454}. More recent work can be found in \cite{5980070,buchli2011learning}. Here, the first one focuses on human-robot interaction, while the second one makes use of RL to learn controller parameters for simple real-world tasks. In \cite{ganesh2010biomimetic} basic concepts from human motor control were transferred to impedance adaptation.

\paragraph{Contribution} Our approach bases on several concepts and connects them in a novel way. Soft robotics \cite{albu2008soft} is considered the basis in terms of hardware and fundamental capabilities, i.e. the embodiment, enabling us in conjunction with impedance control \cite{hogan1984impedance} to apply the idea of learning robot manipulation to rather complex problems. We further extend this by making use of the adaptive impedance controller introduced in \cite{yang2011human}. Both Cartesian stiffness and feed-forward wrench are adapted during execution, depending on the adaptive motion error and based on four interpretable meta parameters per task coordinate. From this follows the question how to choose these meta parameters with respect to the environment and the concrete problem at hand.
To unify all these aspects we introduce a novel robot manipulation skill formalism that acts as a meaningful interpretable connection between problem definition and real world. Its purpose is to reduce the complexity of the solution space of a given manipulation problem by applying a well designed, however still highly flexible structure, see Fig. \ref{fig:complexity}. When applied to a real-world task this structure is supported by an expert skill design and a given quality metric for evaluating the skill's execution. The reduction of complexity is followed by the application of machine learning methods such as CMA-ES \cite{hansen2001completely}, Particle Swarm Optimization \cite{poli2007particle} and Bayesian optimization \cite{shahriari2016taking} to solve the problem not directly on motor control level but in the meta parameter space.

In summary, we put forward the hypothesis that learning manipulation is much more efficient and versatile if we can make use of local intelligence of system components such as the adaptive impedance controller and a well structured skill formalism, essentially encoding abstract expert knowledge. These are the physics grounded computational elements that receive more abstract commands such as goal poses and translate them into basic joint torque behaviors. In this, we draw inspiration from the way humans manipulate their surroundings, i.e. not by consciously determining the muscle force at every time step but rather making use of more complex computational elements \cite{johansson2009coding}.

As a particular real-world example to support our conceptual work, we address the well-known and researched, however, in general still unsolved \cite{kronander2014task} peg-in-hole problem. Especially speed requirements and accuracy still pose significant challenges even when programmed by experts. Different approaches to this problem class were devised. Table \ref{tab:ref_peginhole} depicts a representative selection of works across literature aiming to solve peg-in-hole and categorizes them. As can be seen, insertion times greatly depend on the problem difficulty, although, modern control methodologies are clearly beneficial compared to older approaches. Learning performance has significantly increased over time, however, part of this improvement may have been bought with the need for large computational power e.g. GPUs and computing clusters which might be disadvantageous in more autonomous settings. The difficulty of the considered problem settings in terms of geometry and material varies from industrial standards to much more simple everyday objects.

In summary, our contributions are as follows.
\begin{itemize}
	\item Extension of the adaptive impedance controller from \cite{yang2011human} to Cartesian space and full feed-forward tracking.
	\item A novel meta parameter design for the adaptive controller from \cite{yang2011human} based on suitable real-world constraints of impedance controlled systems.
	\item A novel graph-based skill formalism to describe robot manipulation skills and bridge the gap between high-level specification and low-level adaptive interaction control. Many existing approaches have a more practical and high-level focus on the problem and lack a rigid formulation that is able to directly connect planning with manipulation \cite{pedersen2015robot,thomas2013new,andersen2014definition}.
	\item State-of-the-art learning algorithms such as Covariance Matrix Adaptation \cite{hansen2001completely}, Bayesian optimization \cite{shahriari2016taking} and particle swarm optimization \cite{poli2007particle} are compared experimentally within the proposed framework. 
\item The performance of the proposed framework is showcased for three different (industrially relevant) peg-in-hole problems. We show that the used system can learn tasks complying with industrial speed and accuracy requirements\footnote{In the accompanying video it is demonstrated that on average the used robot is even able to perform the task faster than humans.}.
\item The proposed system is able to learn complex manipulation tasks in a short amount of time of 5-20 minutes depending on the problem while being extremely efficient in terms of raw computational power and memory consumption. In particular, our entire framework can run on a small computer such as the Intel NUC while maintaining a real-time interface to the robot and at the same time running a learning algorithm.
\end{itemize}

The remainder of the paper is organized as follows. Section \ref{sec:controller} describes the adaptive impedance controller, which behavior can be changed fundamentally by adapting its meta parameters. Section \ref{sec:skill} introduces our skill definition and defines the formal problem at hand. In Sec. \ref{sec:learning} the learning algorithms applied to the problem definition are investigated. In Section \ref{sec:experiments} we apply our approach to the well-known peg-in-hole problem. Finally, Sec. \ref{sec:conclusion} concludes the paper.
\section{Adaptive Impedance Controller}\label{sec:controller}
Consider the standard rigid robot dynamics 
\begin{equation}
{M}(\mathbf{q})\ddot{\mathbf{q}}+{C}(\mathbf{q},\dot{\mathbf{q}})\dot{\mathbf{q}}+\mathbf{g}(\mathbf{q})=\bm{\tau}_u+\bm{\tau}_\text{ext},
\end{equation}
where ${M}(\mathbf{q})$ is the symmetric, positive definite mass matrix, ${C}(\mathbf{q},\dot{\mathbf{q}})\dot{\mathbf{q}}$ the Coriolis and centrifugal torques, $\mathbf{g}(\mathbf{q})$ the gravity vector and $\bm{\tau}_{ext}$ the vector of external link-side joint torques. The adaptive impedance control law is defined as
\begin{equation}\label{eq:control_law}
\bm{\tau}_u(t)={J}(\mathbf{q})^T( -\mathbf{F}_{ff}(t)-\V{F}_d(t)-{K}(t)\mathbf{e}-{D}\dot{\mathbf{e}})+\bm{\tau}_r,
\end{equation}
where $\mathbf{F}_{ff}(t)$ denotes the adaptive feed-forward wrench. $\V{F}_d (t)$ is an optional time dependent feed-forward wrench trajectory, ${K}(t)$ the stiffness matrix, $D$ the damping matrix and ${J}(\mathbf{q})$ the Jacobian.
The position and velocity error are $\mathbf{e}=\V{x}^\star - \V{x}$ and $\dot{\mathbf{e}}=\dot{\V{x}}^\star - \dot{\V{x}}$, respectively. ''${}^\star$" denotes the desired motion command.
The dynamics compensator $\bm{\tau}_r$ can be defined in multiple ways, see for example \cite{yang2011human}.
The adaptive tracking error \cite{slotine1991applied} is 
\begin{equation}
\mathbf{\epsilon}=\mathbf{e}+\kappa\dot{\mathbf{e}} ,
\end{equation}
with $\kappa>0$. The adaptive feed-forward wrench $\V{F}_{ff}$ and stiffness $K$ are
\begin{equation}
\V{F}_{ff} = \int_{0}^t  \dot{\boldsymbol{F}}_{ff}(t) \mathrm{d}t + \V{F}_{ff,0},\; K(t)=\int_0^t \dot{K}(t) dt + K_0,
\end{equation}
where $\V{F}_{ff,0}$ and $K_0$ denote the initial values. The controller adapts feed-forward wrench and stiffness by
\begin{align}
\dot{\boldsymbol{F}}_{ff}(t)&=\frac{1}{T}{\alpha}(\boldsymbol{\epsilon}-{\gamma}_\alpha(t)\boldsymbol{F}_{ff}(t)),\label{eq:adapt_ff}\\
\dot{K}(t)&=\frac{1}{T}{\beta}(\text{diag}(\boldsymbol{\epsilon}\circ \boldsymbol{e})-{\gamma}_\beta(t){K}(t)). \label{eq:adapt_k}
\end{align}

The positive definite matrices $\alpha$, $\beta$, $\gamma_\alpha$, $\gamma_\beta$ are the learning rates for feed-forward commands, stiffness and the respective forgetting factors. The learning rates $\alpha$ and $\beta$ determine stiffness and feed-forward adaptation speed. $\gamma_\alpha$ and $\gamma_\beta$ are responsible for slowing down the adaptation process, which is the main dynamical process for low errors. Cartesian damping ${D}$ is designed according to \cite{Albu-SchafferOttFreHir2003} and $T$ denotes the sample time of the controller.
Reordering and inserting \eqref{eq:adapt_ff} and \eqref{eq:adapt_k} into \eqref{eq:control_law} leads to the overall control policy
\begin{equation}
\bm{\tau}_u = f_c (\dot{\mathbf{x}}_d,\mathbf{F}_{d},\alpha,\beta,\gamma_{\alpha},\gamma_{\beta},\mathbf{\Omega}).
\end{equation}
$\mathbf{\Omega}$ denotes the percept vector containing the current pose, velocity, forces etc.

\subsection{Meta Parameter Constraints}
In order to constrain the subsequent meta learning problem we make use of the following reasonable constraints that characterize essentially every physical real-world system. For better readability we discuss the scalar case, which generalizes trivially to the multi-dimensional one. The first constraint of an adaptive impedance controller is the upper bound of stiffness adaptation speed
\begin{equation}
\dot{K}_\text{max}= \max_{t>0} \frac{1}{T}\left[ \beta(\epsilon(t)e(t)- \gamma_\beta K(t)) \right].\label{eq:kdot_max}
\end{equation}
If we now assume that $K(t=0)=0$ and $\dot{e}=0$ we may define $e_{max}$ as the error at which $\dot{K}_{max}=\frac{\beta e^2_{max}}{T}$ holds. Then, the maximum value for $\beta$ can be written as
\begin{equation}
\beta_{\mathrm{max}}=\frac{\dot{K}_{\mathrm{max}}T}{e_{\mathrm{max}}^2}.\label{eq:beta_max}
\end{equation}
Furthermore, the maximum decrease of stiffness occurs for $\V{e}=\V{0}$ and $K(t)=K_\text{max}$, where $K_\text{max}$ denotes the maximum stiffness, also being an important constraint for any real-world impedance controlled robot. Thus, we may calculate an upper bound for $\gamma_\beta$ as
\begin{equation}
\gamma_{\beta\text{,max}} = \dfrac{\dot{K}_\text{max}}{\beta_\text{max} K_\text{max}}.
\end{equation}

Finding the constraints of the adaptive feed-forward wrench may be done analogously. In conclusion, we relate the upper limits for $\alpha$, $\beta$, $\gamma_\alpha$ and $\gamma_\beta$ to the inherent system constraints $K_{max}$, $F_{max}$, $\dot{K}_{max}$ and $\dot{F}_{max}$.
\section{Manipulation Skill}\label{sec:skill}
In this section we introduce a mathematical formalism to describe robot manipulation skills. A manipulation skill is defined as the directed graph $G$ consisting of nodes $n \in N$ and edges $e \in E$. A node $n \in N$ is also called a manipulation primitive (MP), while an edge $e \in E$ is also called transition. The transitions are activated by conditions that mark the success of the preceding MP. A single MP consists of a parameterized twist and feed forward wrench trajectory 

\begin{align*}
\dot{\mathbf{x}}_{d} &= f_{t}(\mathbf{P}_{t},\V{\Omega}),\\
\mathbf{F}_{d} &= f_{w}(\mathbf{P}_{w},\V{\Omega}),
\end{align*}

where $\dot{\mathbf{x}}_{d}$ is the desired twist and $\mathbf{F}_{d}$ the feed forward wrench skill command. These commands are described with respect to a task frame $TF$.
The percept vector $\V{\Omega}$ is required since information about the current pose or external forces may in general be needed by the MPs.
$\mathbf{P}_{t}$ is the set of parameters used by node $n$ to generate twist commands while $\mathbf{P}_{w}$ is used to generate the wrench commands. Moreover, $\mathbf{P}_{t} \subset \mathcal{P}$ and $\mathbf{P}_{w} \subset \mathcal{P}$, where $\mathcal{P}$ is the set of all parameters.
Furthermore, we divide $\mathcal{P}$ into two different subsets $\mathcal{P}_c$ and $\mathcal{P}_l$. The parameters in $\mathcal{P}_c$ are entirely determined by the context of the task e.g. geometric properties of involved objects. $\mathcal{P}_l$ is the subset of all parameters that are to be learned. They are chosen from a domain $\mathcal{D}$ which is also determined by the task context or system capabilities.

\begin{figure}[ht!]
\begin{center}
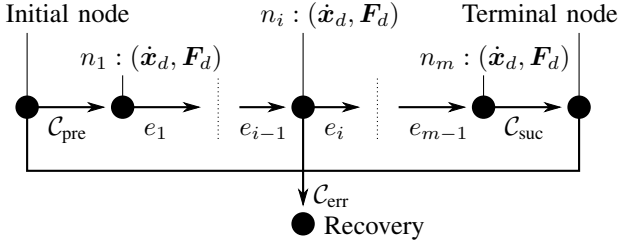
\caption{Skill graph $G$.}
\label{fig:skill}
\end{center}
\vspace{-0.4cm}
\end{figure}

Figure \ref{fig:skill} shows a principle depiction of the graph $G$ with an additional initial and terminal node. The transition coming from the initial node is triggered by the \textit{precondition} while the transition leading to the terminal node is activated by the \textit{success condition}. Furthermore, every MP has an additional transition that is activated by an \textit{error condition} and leads to a recovery node.
When a transition is activated and the active node switches from $n_i$ to $n_{i+1}$ the switching behavior should be regulated.

Note that, although it is possible, we do not consider backwards directed transitions in this work since this would introduce another layer of complexity to the subsequent learning problem that is out of the scope of this paper. Clearly, this would rather become a high-level planning problem that requires more sophisticated and abstract knowledge.

\subsubsection*{Conditions}

There exist three condition types involved in the execution of skills: preconditions $\mathcal{C}_\text{pre}$, error conditions $\mathcal{C}_\text{err}$ and success conditions $\mathcal{C}_\text{suc}$. They all share the same basic definition, yet their application is substantially different. In particular, their purpose is to define the limits of the skill from start to end.
The precondition states the conditions under which the skill can be initialized. The error condition stops the execution when fulfilled and returns a negative result. The success condition also stops the skill and returns positive. Note, that we also make use of the success condition definition to describe the transitions between MPs.

\begin{definition}[Condition]
Let $\mathcal{C} \subset \mathsf{S}$ be a closed set and $\mathsf{c}(\mathbf{X}(t))$ a function $\mathsf{c}: \mathcal{S}\rightarrow \mathbb{B}$ where $\mathbb{B}=\{0,1\}$. A condition holds iff $\mathsf{c}(\mathbf{X}(t))=1$. Note that the mapping itself obviously depends on the specific definition type.
\end{definition}

\begin{definition}[Precondition]
$\mathcal{C}_{pre}$ denotes the chosen set for which the precondition defined by $\mathsf{c}_{pre}(\mathbf{X}(t))$ holds. The condition holds, i.e. $\mathsf{c}_{pre}(\mathbf{X}(t_0))=1$, iff $\forall \; x \in \mathbf{X}: x(t_0) \in \mathcal{C}_{pre} $. $t_0$ denotes the time at start of the skill execution.
\end{definition}

\begin{definition}[Error Condition]
$\mathcal{C}_{err}$ denotes the chosen set for which the error condition $\mathsf{c}_{err}(\mathbf{X}(t))$ holds, i.e. $\mathsf{c}_{err}(\mathbf{X}(t))=1$. This follows from $\exists \; x \in \mathbf{X}: x(t) \in \mathcal{C}_{err} $.
\end{definition}

\begin{definition}[Success Condition]
$\mathcal{C}_{suc}$ denotes the chosen set for which the success condition defined by $\mathsf{c}_{suc}(\mathbf{X}(t))$ holds, i.e. $\mathsf{c}_{suc}(\mathbf{X}(t))=1$ iff $\forall x \in \mathbf{X}: x(t) \in \mathcal{C}_{suc} $.
\end{definition}

\subsubsection*{Evaluation}

Lastly, a learning metric is used to evaluate the skill execution in terms of a success indicator and a predefined cost function.

\begin{definition}[Learning Metric]
$\mathcal{Q}$ denotes the set of all $2-$tuples $(w,f_q(\mathbf{X}(t))$ with $0<w<1$ and the result indicator $r=\{0,1\}$ where $0$ denotes failure and $1$ success. 
Let $q=\sum_i w_i f_{q,i}(\mathbf{X}(t)) \;\forall \; (w_i,f_{q,i}(\mathbf{X}(t))) \in \mathcal{Q}$ be the cost function of the skill.
\end{definition}

Note that the learning metric is designed according to the task context and potential process requirements. Examples for the peg-in-hole skill would be the insertion time or the average contact forces during insertion.

There exist other relevant works that make use of manipulation primitives such as \cite{finkemeyer2005executing,kroger2010manipulation} in which MPs are used to switch between specific situation dependent control and motion schemes. They are composed to MP nets to enable task planning. Similar approaches can be found in \cite{weidauer2014hierarchical,pek2016simplifying}.

\subsection{Peg-in-Hole}\label{sec:spec_peginhole}

\begin{figure*}
\begin{center}
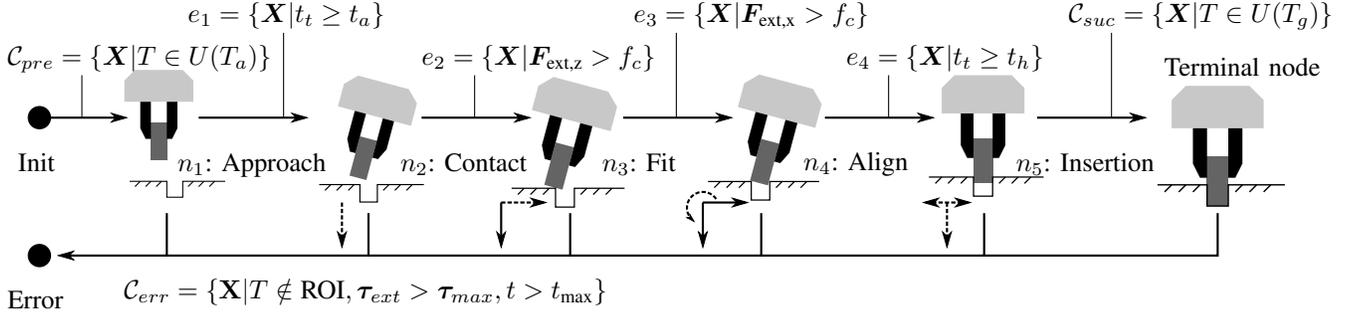
\caption{Visualization of the peg-in-hole skill graph. Dashed arrows denote velocity commands, solid ones feed forward force commands. $t_1$ and $t_4$ are the trajectory durations in states $n_1$ and $n_4$.}
\label{fig:peginhole}
\end{center}
\vspace{-0.6cm}
\end{figure*}

In the following, the well-known, however, still challenging peg-in-hole skill is described with the help of the above formalism. Figure \ref{fig:peginhole} shows the graph of the skill including the manipulation primitives. The parameters $p \in \mathcal{P}$ are the estimated hole pose $T_\text{h}$, the region of interest around the hole ROI, the depth of the hole $d$, the maximum allowed velocity and acceleration for translation and rotation $\dot{\mathbf{x}}_\text{max}$ and $\ddot{\mathbf{x}}_\text{max}$, the initial tilt of the object relative to the hole $\varphi_\text{init}$, the force $f_\text{c}$, the speed factor $s$, the amplitude of translational and rotational oscillations $a_t$, $a_r$ and their frequencies $\omega_t$, $\omega_r$.


\subsubsection*{Manipulation Primitives}

The MPs for peg-in-hole are defined as follows:
\begin{itemize}
\item $n_1$: The robot moves to the approach pose.
\begin{align*}
\dot{\mathbf{x}}_{d}=f_\text{p2p}(T_\text{a},s \dot{\mathbf{x}}_\text{max},\ddot{x}_\text{max},),\;
\mathbf{F}_{d}=\mathbf{0}.
\end{align*}
$f_\text{p2p}$ generates a trapezoidal velocity profile to move from the current pose to a goal pose while considering a given velocity and acceleration.
\item $n_2$: The robot moves towards the surface with the hole and establishes contact.
\begin{align*}
\dot{\mathbf{x}}_{d}=\left[\begin{array}{cccccc}
0 & 0 & s \dot{\mathbf{x}}_\text{max} & 0 & 0 & 0
\end{array}\right]^T,\;
\mathbf{F}_{d}=\mathbf{0}.
\end{align*}
\item $n_3$: The object is moved laterally in $x$-direction until it is constrained.
\begin{align*}
\dot{\mathbf{x}}_{d}&=\left[\begin{array}{cccccc}
s \dot{\mathbf{x}}_\text{max} & 0 & 0 & 0 & 0 & 0
\end{array}\right]^T,\\
\mathbf{F}_{d}&=\left[\begin{array}{cccccc}
0 & 0 & f_\text{c} & 0 & 0 & 0
\end{array}\right]^T.
\end{align*}
\item $n_4$: The object is rotated into the estimated orientation of the hole while keeping contact with the hole.
\begin{align*}
\dot{\mathbf{x}}_{d}&=f_\text{p2p}(T_\text{h},s \dot{\mathbf{x}}_\text{max},\ddot{\mathbf{x}}_\text{max}), \\
\mathbf{F}_{d}&=\left[\begin{array}{cccccc}
f_\text{c} & 0 & f_\text{c} & 0 & 0 & 0
\end{array}\right]^T.
\end{align*}
\item $n_5$: The object is inserted into the hole.
\begin{align*}
\dot{\mathbf{x}}_{d}=\left[\begin{array}{c}
a_t \sin\left(2\pi \omega_t\right) \\ a_t \sin\left(2\pi \frac{3}{4}\omega_t\right)\\ s \dot{\mathbf{x}}_\text{max} \\ a_r \sin\left(2\pi \omega_r\right) \\ a_r \sin\left(2\pi \frac{3}{4}\omega_r\right) \\ 0
\end{array}\right]^T, \;\mathbf{F}_{d}=\mathbf{0}
\end{align*}
\end{itemize}
\section{Parameter Learning}\label{sec:learning}
\begin{figure*}
\begin{center}
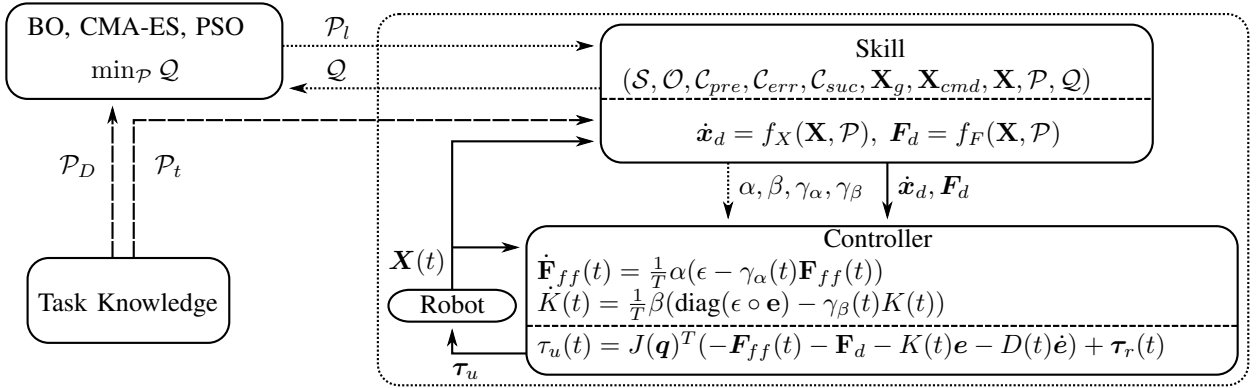
\caption{Implementation of skill learning framework. Solid lines (---) indicate a continuous flow of information, dotted lines ($\cdot\cdot\cdot$) relate to information exchange only once per trial and dashed lines $(---)$ to information relayed only once per experiment.}
\label{fig:learning_overview}
\end{center}
\vspace{-0.8cm}
\end{figure*}

Figure \ref{fig:learning_overview} shows how the controller and the skill formalism introduced in Sec. \ref{sec:controller} and \ref{sec:skill} are connected to a given learning method to approach the problem of meta learning, i.e., finding the right (optimal) parameters for solving a given task. The execution of a particular learning algorithm until it terminates is named an experiment throughout the paper. A single evaluation of parameters is called a trial.

\subsection{Requirements}

A potential learning algorithm will be applied to the system defined in Sections \ref{sec:controller} and \ref{sec:skill}. In particular, it has to be able to evaluate sets of parameters per trial in a continuous experiment. Since we apply it to a real-world physical manipulation problem the algorithm will face various challenges that result in specific requirements.

\begin{enumerate}
\item Generally, no feasible analytic solution
\item Gradients are usually not available
\item Real world problems are inherently stochastic
\item No assumptions possible on minima or cost function convexity
\item Violation of safety, task or quality constraints
\item Significant process noise and many repetitions
\item Low desired total learning time
\end{enumerate}

Thus, suitable learning algorithms will have to fulfill subsequent requirements. They must provide a numerical black-box optimization and cannot rely on gradients. Stochasticity must be regarded and the method has to be able to optimize globally. Furthermore, it should handle unknown and noisy constraints and must provide fast convergence rates.

\subsection{Comparison}
Table \ref{tab:opt_methods} lists several groups of state-of-the-art optimization methods and compares them with respect to above requirements. In this, we follow and extend the reasoning introduced in \cite{calandra2014experimental}. It also shows that most algorithms do not fulfill the necessary requirements. Note that for all algorithms there exist extensions for the stochastic case. However, comparing all of them is certainly out of the scope of the paper. Therefore, we focus on the most classical representatives of the mentioned classes.

\renewcommand{\arraystretch}{1.3}
\begin{table}[htp!]
\begin{center}
\caption{Suitability of existing learning algorithms with regard to the properties no gradient (NG), stochasticity assumption (SA), global optimizer (GO) and unknown constraints (UC).}
\label{tab:opt_methods}
\begin{tabular}{ccccccc}
\hline 
Method & NG & SA & GO & UC \\
\hline 
Grid Search & $+$ & $-$ & $+$ & $-$  \\ 
\hline 
Pure Random Search & $+$ & $-$ & $+$ & $-$ \\ 
\hline 
Gradient-descent family & $-$ & $-$ & $-$ & $-$ \\ 
\hline 
Evolutionary Algorithms & $+$ & $-$ & $+$ & $-$ \\ 
\hline 
Particle Swarm & $+$ & $+$ & $+$ & $-$ \\ 
\hline 
Bayesian Optimization & $+$ & $+$ & $+$ & $+$ \\ 
\end{tabular}
\end{center}
\vspace{-0.4cm}
\end{table}

Generally, \emph{gradient-descent} based algorithms require a gradient to be available, which obviously makes this class unsuitable. \emph{Grid search}, pure \emph{random search} and \emph{evolutionary algorithms} typically do not assume stochasticity and cannot handle unknown constraints very well without extensive knowledge about the problem they optimize, i.e. make use of well-informed barrier functions. The latter aspect applies also to \emph{particle swarm} algorithms. Only Bayesian optimization (BO) in accordance to \cite{snoek2013bayesian} is capable of explicitly handling unknown noisy constraints during optimization.

Although it seems that Bayesian optimization is the most suited method to cope with our problem definition some of the other algorithm classes might also be capable of finding solutions, maybe even faster. Thus, in addition, we select Covariance Matrix Adaptation Evolutionary Strategy (CMA-ES) \cite{hansen2001completely} and Particle Swarm Optimization (PSO) \cite{poli2007particle} for comparison. Furthermore, we utilize Latin Hypercube Sampling (LHS) \cite{mckay1979comparison} (an uninformed sampler) to gain insights into the difficulty of the problems at hand.

\subsubsection{Bayesian Optimization}
For Bayesian optimization we made use of the spearmint software package \cite{snoek2013bayesian,snoek2012practical,brochu2010tutorial,swersky2013multi}. In general, BO finds the minimum of an unknown objective function $f(\V{p})$ on some bounded set $\mathcal{X}$ by developing a statistical model of $f(\boldsymbol{p})$. Apart from the cost function, it has two major components, which are the prior and the acquisition function.
\begin{itemize}
\item Prior: We use a Gaussian process as prior to derive assumptions about the function being optimized. The Gaussian process has a mean function $m: \mathcal{X} \rightarrow \mathbb{R}§$ and a covariance function $B: \mathcal{X} \times \mathcal{X} \rightarrow \mathbb{R}$. As a kernel we use the automatic relevance determination (ARD) Mat\'{e}rn $5/2$ kernel
\begin{align*}
B_{M52}(\V{p},\V{p}^\prime)=&\theta_0(1+\sqrt{5 r^2(\V{p},\V{p}^\prime)}\\
&+\frac{5}{3}r^2(\V{p},\V{p}^\prime) e^{-\sqrt{5 r^2(\V{p},\V{p}^\prime)}},
\end{align*}
with
\begin{equation*}
r^2(\V{p},\V{p}^\prime)=\sum_{i=1}^d \frac{(p_i-p_i^\prime)^2}{\theta_i^2}.
\end{equation*}
This kernels results in twice-differentiable sample functions which makes it suitable for practical optimization problems as stated in \cite{snoek2012practical}. 
It has $d+3$ hyperparameters in $d$ dimensions, i.e. one characteristic length scale per dimension, the covariance amplitude $\theta_0$, the observation noise $\nu$ and a constant mean $m$. These kernel hyperparameters are integrated out by applying Markov chain Monte Carlo (MCMC) via slice sampling \cite{neal2003slice}.
\item Acquisition function: We use predictive entropy search with constraints (PESC) as a means to select the next parameters $\boldsymbol{p}$ to explore, as described in \cite{hernandez2015predictive}.
\end{itemize}

\subsubsection{Latin Hypercube Sampling}

Latin hypercube sampling (LHS) \cite{mckay1979comparison} is a method to sample a given parameter space in a nearly random way. In contrast to pure random sampling LHS generates equally distributed random points in the parameter space. It might indicate whether complexity reduction of the manipulation task was successful when it is possible to achieve feasible solutions by sampling.

\subsubsection{Covariance Matrix Adaptation}

The Covariance Matrix Adaptation Evolutionary Strategy (CMA-ES) is an optimization algorithm from the class of evolutionary algorithms for continuous, non-linear, non-convex black-box optimization problems \cite{hansen2001completely,hansen2006cma}.

The algorithm starts with an initial centroid $m \in \mathfrak{R}^n$, a population size $\lambda$, an initial step-size $\sigma>0$, an initial covariance matrix $C=I$ and isotropic and anisotropic evolution paths $p_\sigma=0$ and $p_c=0$. $m$, $\lambda$ and $\sigma$ are chosen by the user. Then the following steps are executed until the algorithm terminates.
\begin{enumerate}
\item Evaluation of $\lambda$ individuals sampled from a normal distribution with mean $m$ and covariance matrix $\sigma C$.
\item Update of centroid $m$, evolution paths $p_\sigma$ and $p_c$, covariance matrix $C$ and step-size $\sigma$ based on the evaluated fitness.
\end{enumerate}

\subsubsection{Particle Swarm Optimization}

Particle swarm optimization usually starts by initializing all particle's positions $\V{x}_i(0)$ and velocities $\V{v}_i(0)$ with a uniformly distributed random vector, i.e. $\V{x}_i(0) \sim  U(\V{b}_\text{lb},\V{b}_\text{ub})$ and $\V{v}_i(0) \sim  U(-\vert \V{b}_\text{ub} - \V{b}_\text{lb} \vert,\vert \V{b}_\text{ub} - \V{b}_\text{lb} \vert)$ with $U$ being the uniform distribution. The particles are evaluated at their initial positions and their personal best $\V{p}_i$ and the global best $\V{g}$ are set.
Then, until a termination criterion is met, following steps are executed:
\begin{enumerate}
\item Update particle velocity:
\begin{align*}
\V{v}_i(t+1)=&\V{v}_i(t)+c_1(\V{p}_i-\V{x}_i(t))R_1\\
&+\V{v}_i(t)+c_2(\V{g}-\V{x}_i(t))R_2
\end{align*}
where $R_1$ and $R_2$ are diagonal matrices with random numbers generated from a uniform distribution in $[0,1]$ and $c_1,c_2$ are acceleration constants usually in the range of $[0,4]$.
\item Update the particle position:
\begin{equation}
\V{x}_i(t+1)=\V{x}_i(t)+\V{v}_i(t+1)
\end{equation}
\item Evaluate the fitness of the particle $f(\V{x}_i(t+1))$.
\item Update each $\V{p}_i$ and global best $\V{g}$ if necessary.
\end{enumerate}

\section{Experiments}\label{sec:experiments}
In our experiments we investigate the learning methods selected in Sec.  \ref{sec:learning} and compare them for three different peg-in-hole variations for the introduced skill formalism and controller, see Sections \ref{sec:controller} and \ref{sec:skill}. The experimental setup consists of a Franka Emika Panda robot \cite{haddadin2017franka} that executes the following routine:
\begin{enumerate}
\item The robot grasps the object to be inserted.
\item A human supervisor teaches the hole position which is fixed with respect to the robot. The teaching accuracy was below $1$ mm.
\item A learning algorithm is selected and executed until it terminates after a specified number of trials.
\item For every trial the robot evaluates the chosen parameters with four slightly different (depending on the actual problem) initial positions in order to achieve a more robust result.
\end{enumerate}

\begin{figure}
\begin{center}
\begin{subfigure}[c]{0.2\textwidth}
\includegraphics[scale=0.02]{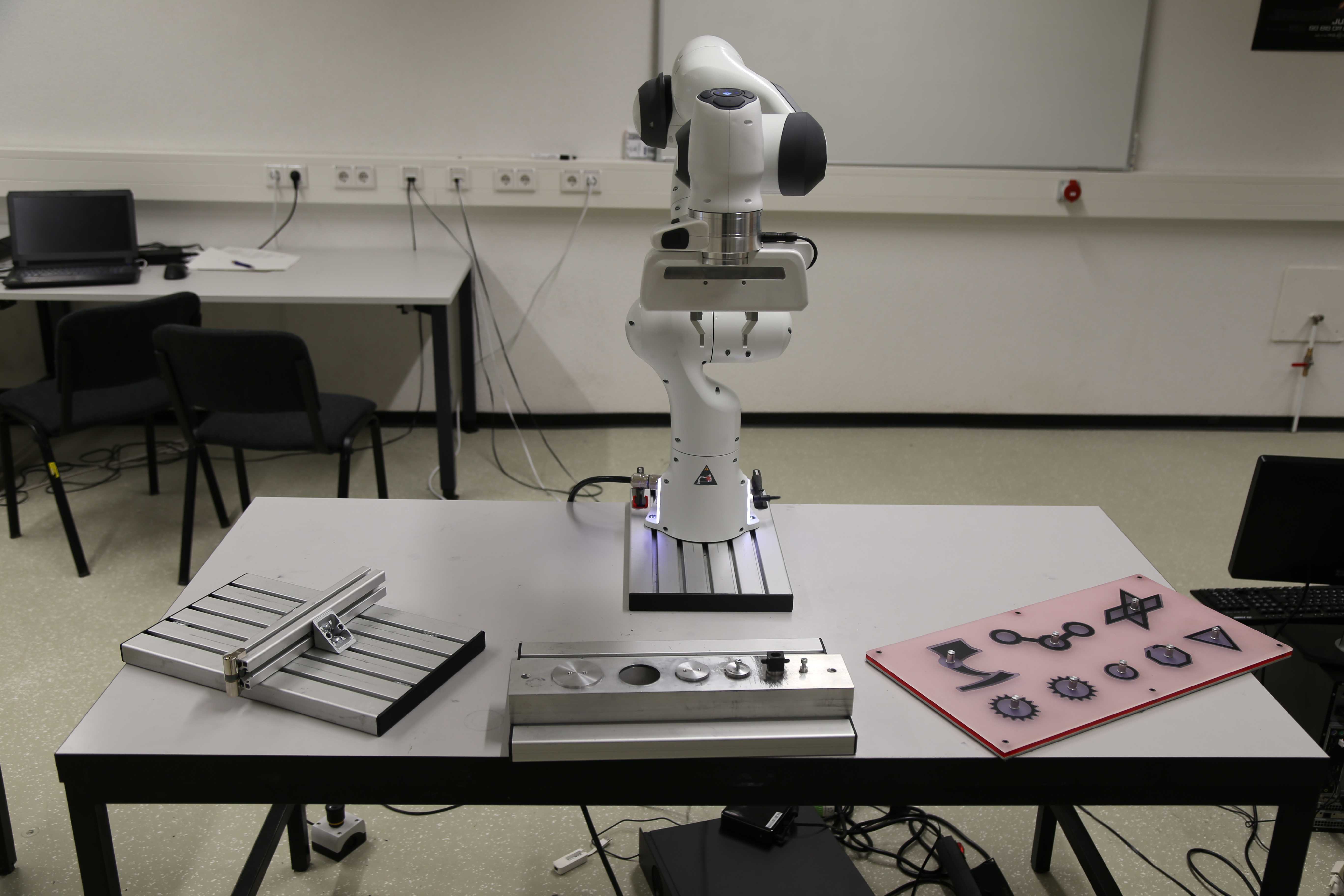}
\end{subfigure}
\begin{subfigure}[c]{0.2\textwidth}
\includegraphics[scale=0.02]{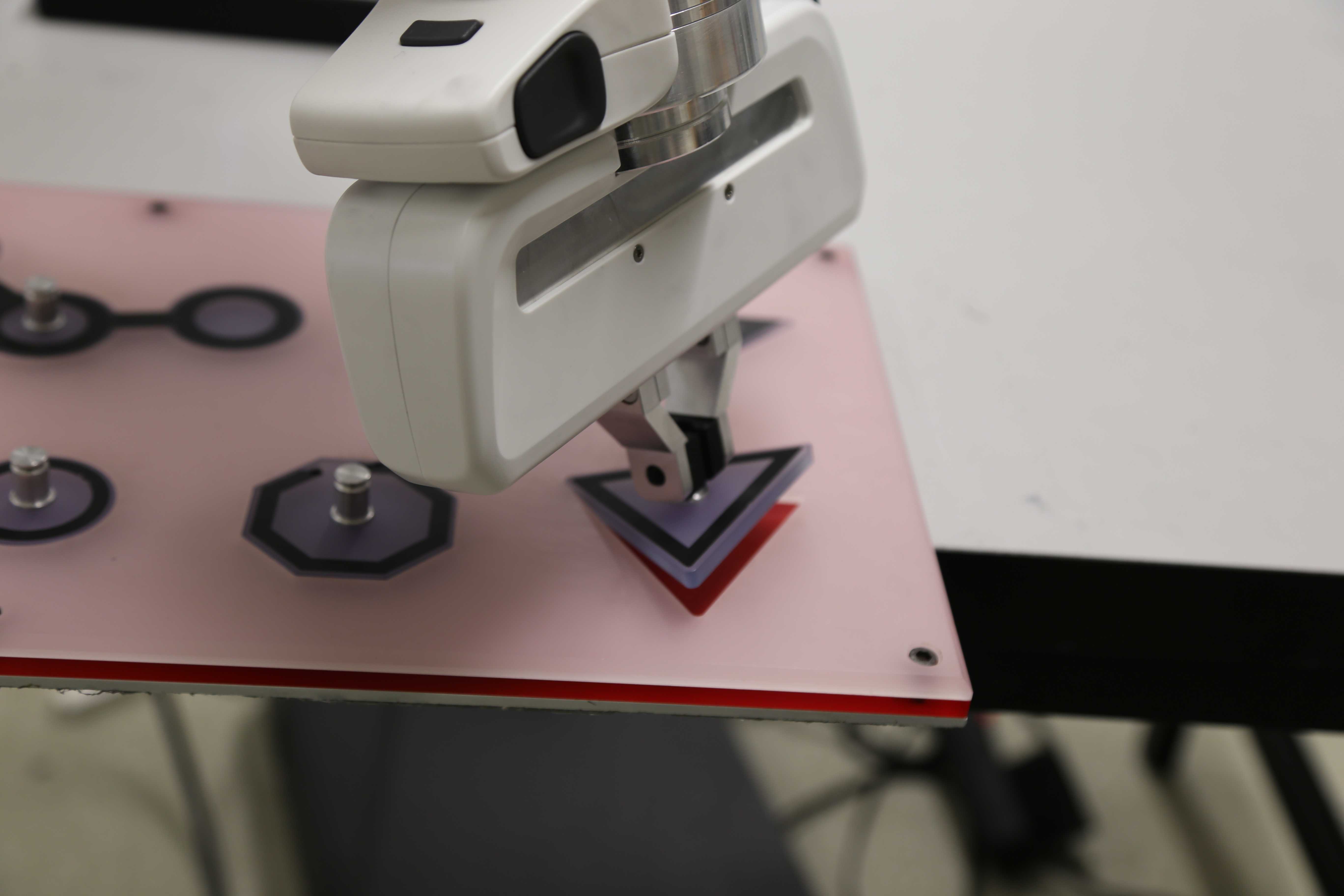}
\end{subfigure}\\
\begin{subfigure}[c]{0.2\textwidth}
\includegraphics[scale=0.02]{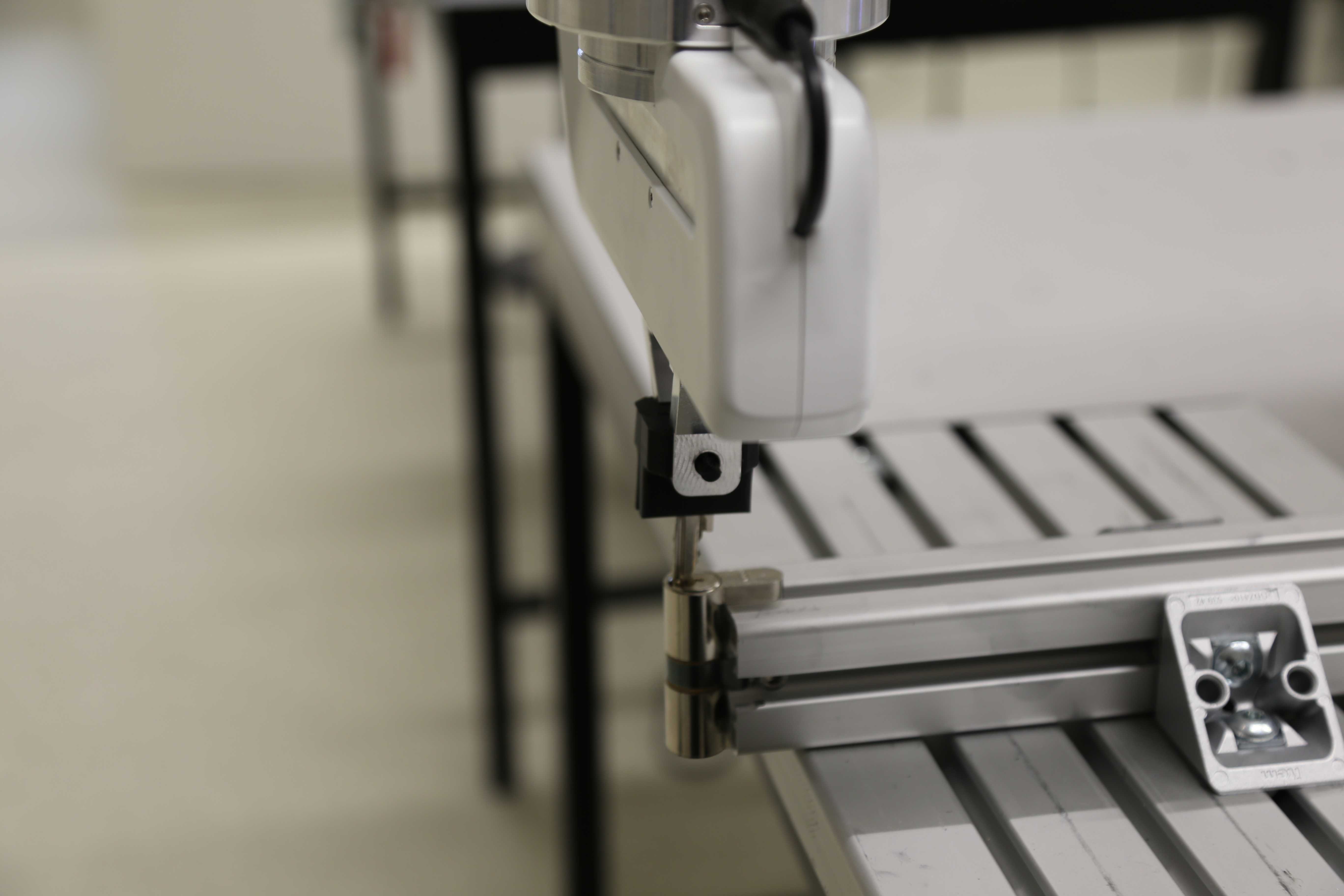}
\end{subfigure}
\begin{subfigure}[c]{0.2\textwidth}
\includegraphics[scale=0.02]{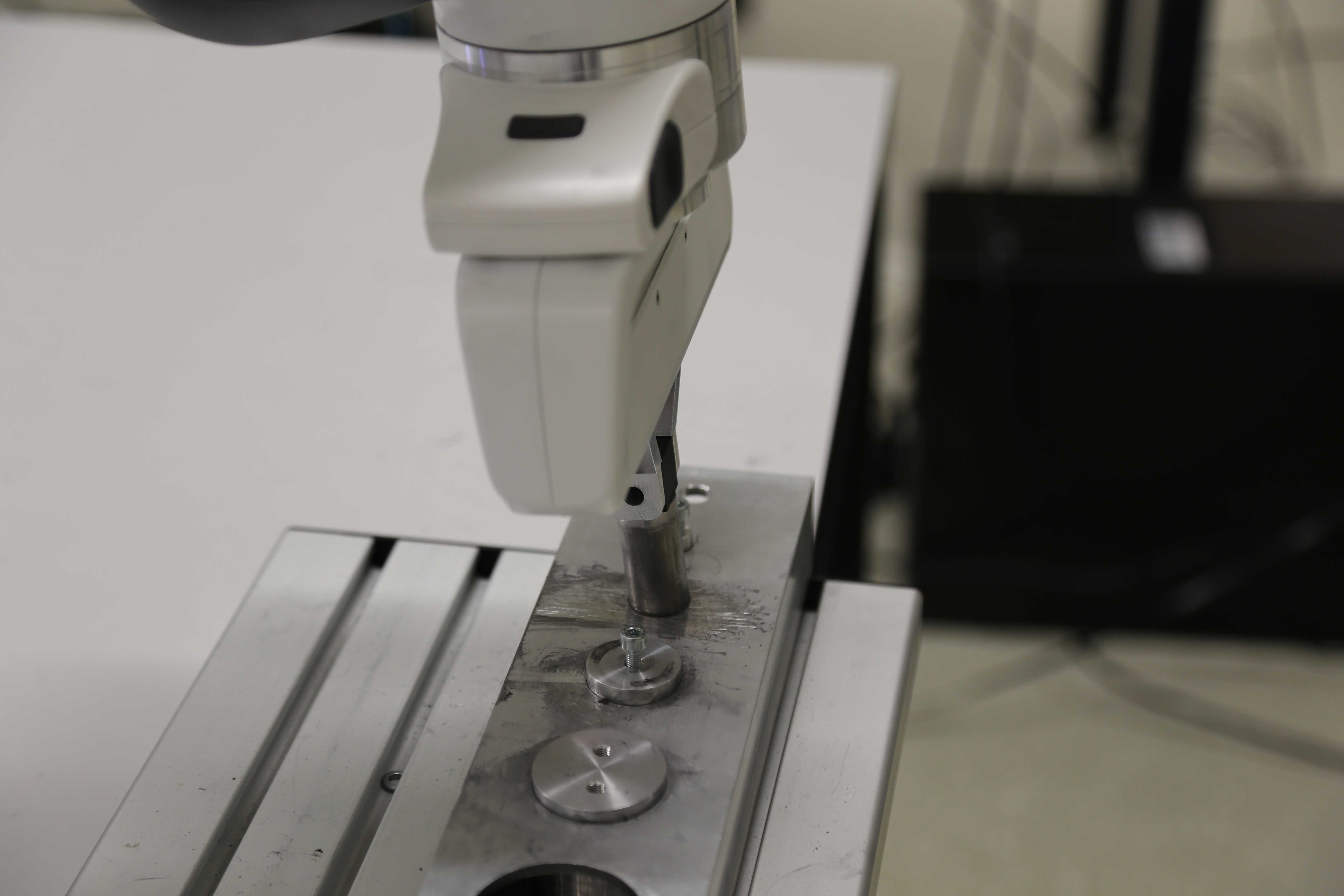}
\end{subfigure}
\end{center}
\caption{Experimental setup, puzzle (top right), key (bottom left) and peg (bottom right).}
\label{fig:exp_setup}
\vspace{-0.4cm}
\end{figure}

We investigated three variations of peg-in-hole as shown in Fig. \ref{fig:exp_setup}, a key, a puzzle piece and an aluminum peg. The meta-parameters and skill parameters were learned with the four methods introduced in Sec. \ref{sec:learning}.

The parameter domain (see Tab. \ref{tab:params_domain}) is the same for most of the parameters that are learned in the experiments and can be derived from system and safety limits which are the maximum stiffness $K_\text{max}=\left[\begin{array}{cc}
2000 \text{ N/m}& 200 \text{ Nm/rad}
\end{array}\right] $, the maximum stiffness adaptation speed $\dot{K}_\text{max}=\left[\begin{array}{cc}
5000 \text{ N/ms}& 500 \text{ Nm/rads}
\end{array}\right] $, the maximum allowed feed forward wrench and wrench adaptation of the controller $\V{F}_\text{ff}=\left[\begin{array}{cc}
10 \text{ N}& 5 \text{ Nm}
\end{array}\right] $ and $\dot{\V{F}}_\text{ff}=\left[\begin{array}{cc}
1 \text{ N/s}& 0.5 \text{ Nm/s}
\end{array}\right]$, the maximum error $e_\text{max}=\left[\begin{array}{cc}
0.005 \text{ m} & 0.017 \text{ rad}
\end{array}\right]$ and the maximum velocity $\dot{\V{x}}_\text{max}=\left[\begin{array}{cc}
0.1 \text{ m/s}& 1 \text{ rad/s}
\end{array}\right]$.

The domain of the learned parameters $\mathcal{P}_l$ is derived from these limits and shown in Tab. \ref{tab:params_domain}.

\renewcommand{\arraystretch}{1.3}
\begin{table}
\begin{center}
\caption{Parameter domain.}
\label{tab:params_domain}
\begin{tabular}{|c|c|c|}
\hline
Parameter & Min & Max \\
\hline
$\alpha$ & $\left[\begin{array}{cc}0 & 0 \end{array}\right]$ & $\left[\begin{array}{cc}0.4 & 1.1765 \end{array}\right]$ \\
\hline
$\beta$ & $\left[\begin{array}{cc}0 & 0 \end{array}\right]$ & $\left[\begin{array}{cc}200000 & 173010 \end{array}\right]$\\
\hline
$\gamma_\alpha$ & $\left[\begin{array}{cc}0 & 0 \end{array}\right]$ & $\left[\begin{array}{cc}5e-4 & 3.4e-4 \end{array}\right]$\\
\hline
$\gamma_\beta$ & $\left[\begin{array}{cc}0 & 0 \end{array}\right]$ & $\left[\begin{array}{cc}0 & 0 \end{array}\right]$\\
\hline
$F_c$ & $5$ N & $15$ N\\
\hline
$s$ & $0$ & $1$ \\
\hline
$a_t$ & $0$ m & $0.005$ m \\
\hline
$\omega_t$ & $0$ Hz & $3.2$ Hz \\
\hline
$\omega_r$ & $0$ Hz & $4.5$ Hz \\
\hline
$\varphi_\text{init}$ & $0$ rad & $0.349$ rad \\
\hline
\end{tabular}
\end{center}
\vspace{-0.8cm}
\end{table}

In the following the specifics of the three tasks shown in Fig. \ref{fig:exp_setup} are explained.
\begin{itemize}
\item Puzzle: The puzzle piece is an equilateral triangle with a side length of $0.075$ m.
The maximum rotational amplitude of the oscillatory motion is given by $a_r=0.09$ rad. The hole has a depth of $d=0.005$ m. The tolerances between puzzle piece and hole are $< 0.1$ mm and there are no chamfers.
\item Key: The key has a depth of $d=0.0023$ m. Since the hole and the key itself have chamfers to make the initial insertion very easy we omit the learning of the initial alignment $\varphi_\text{init}$ and set it to $\varphi_\text{init}=0$ rad. The maximum rotational amplitude of the oscillatory motion is given by $a_r=0.0175$ rad. Due to the very small hole no deliberate initial pose deviation was applied.
\item Peg: The aluminum peg has a diameter of $0.02$ m and the hole has a depth of $0.035$ m. The tolerances are $<< 0.1$ mm and there is a $0.5$ mm chamfer on the peg. The maximum rotational amplitude of the oscillatory motion is given by $a_r=0.035$ rad. The hole has no walls which results in a higher chance of getting stuck during insertion further increasing the difficulty.
\end{itemize}

The learning algorithms are configured as follows.

\begin{itemize}
\item LHS: The parameter space is sampled at $75$ points.
\item CMA-ES: The algorithm ran for $15$ generations with a population of $5$ individuals and $\sigma_0=0.1$. The initial centroid was set in the middle of the parameter space.
\item PSO: We used $25$ particles and let the algorithm run for $3$ episodes. The acceleration constants were set to $c_1=2$ and $c_2=2$.
\item BO: The algorithm is initialized with $5$ equally distributed random samples from the parameter space.
\end{itemize}

As cost function for all three problems we used the execution time measured from first contact to full insertion. A maximum execution time of $15$ s was part of the skill's error condition. In case the error condition was triggered we added the achieved distance towards the goal pose in the hole to the maximum time multiplied by a factor.

\subsection{Results}

\begin{figure*}[ht!]
\begin{center}
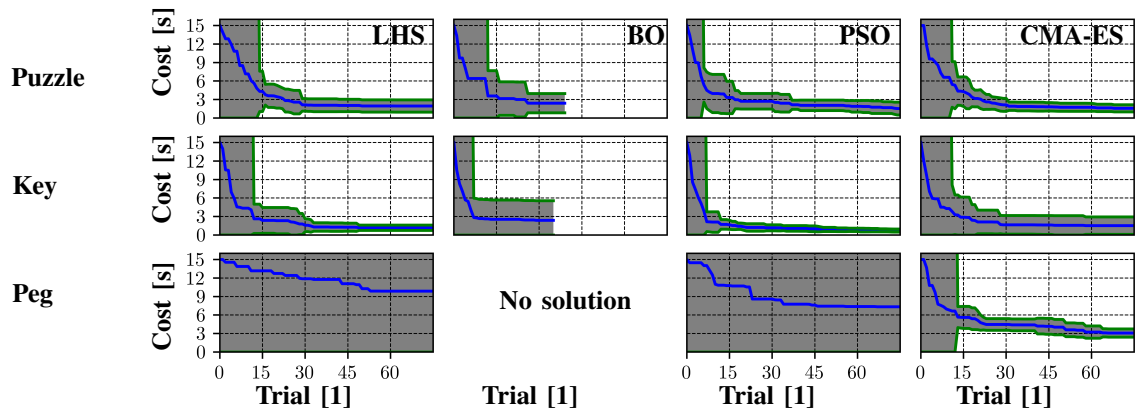
\caption{Experimental results. Columns show the plots for the learning algorithms (LHS,BO,PSO,CMA-ES) and rows for the tasks (Puzzle,Key,Peg).}
\label{fig:plots}
\end{center}
\vspace{-0.8cm}
\end{figure*}

The results can be seen in Fig. \ref{fig:plots}. The blue line is the mean of the execution time per trial averaged over all experiments. The grey area indicates the $90$ \% confidence interval.

The results indicate that all four algorithms are suited to a certain degree to learn easier variations of the peg-in-hole task such as the puzzle and the key. However, it must be noted that Bayesian optimization on average finds solutions not as good as the other methods. Furthermore, the confidence interval is notably larger. It also terminates early into the experiment since the model was at some point not able to find further suitable candidates. This might indicate a solution space with high noise and discontinuities that is difficult to model.

The comparison with LHS also indicates that the complexity reduction of our formal approach to manipulation skills makes it possible to find solutions with practically relevant execution times by sampling rather then explicit learning.

At the bottom of Fig. \ref{fig:plots} the results for the most difficult peg-in-hole variation are shown. We do not include the results of BO since it was not able to find a solution in the given time frame. The plot showing the LHS method indicates a very hard problem. Random sampling leads to feasible solutions, however, the confidence interval is too large to conclude any assurance. PSO achieves better solutions yet also has a very high confidence interval. CMA-ES outperforms both methods and is able to find a solution that is better in terms of absolute cost and confidence.

Considering the best performing algorithm CMA-ES a feasible solution for any of the tasks was already found after $2-4$ minutes and optimized after $5-20$ minutes depending on the task, significantly outperforming existing approaches for learning peg-in-hole, see Tab. \ref{tab:ref_peginhole}. Note also that with the exception of BO no noteworthy computation time was necessary.
\section{Conclusion}\label{sec:conclusion}
In this paper we introduced an approach to learning robotic manipulation that is based on a novel skill formalism and meta learning of adaptive controls. Overall, the proposed framework is able to solve highly complex problems such as learning peg-in-hole with sub-millimeter tolerances in a low amount of time, making it even applicable for industrial use and outperforming existing approaches in terms of learning speed and resource demands. Remarkably, the used robot was even able to outperform humans in execution time.

Summarizing the results we conclude that the application of complexity reduction via adaptive impedance control and simple manipulation strategies (skills) in combination with the right learning method to complex manipulation problems is feasible. The results might further indicate that methods supported by a certain degree of random sampling are possibly better suited for learning this type of manipulation skills than those more relying on learning a model of the cost function. These conclusions will be investigated more thoroughly in future work.

Overall, our results show that, in contrast to purely data-driven learning/control approaches that do not make use of task structures nor the concept of (adaptive) impedance, our approach of combining sophisticated adaptive control techniques with state-of-the-art machine learning makes even such complex problems tractable and applicable to the real physical world and its highly demanding requirements. Typically, other existing manipulation learning schemes require orders of magnitude more iterations already in simulation while it is known that nowadays simulations cannot realistically (if at all) capture real-world contacts.

Clearly, the next steps we intend to take are the application of our framework to other manipulation problem classes and the thorough analysis of the generalization capabilities of the system to similar, however, yet unknown problems.
\bibliography{references}

\end{document}